\documentclass[12pt, a4paper]{article}
\usepackage[a4paper,
           rmargin=1in]{geometry}
\usepackage{graphicx}
\usepackage{mathrsfs}
\usepackage{amsfonts}
\usepackage{amsmath}

\begin{document}  

\title{On the utility of feature selection in building two-tier decision trees}

\author{S.A. Saltykov}

\maketitle

\begin{abstract}
Nowadays, feature selection is frequently used in machine learning when there is a risk of performance degradation due to overfitting or when computational resources are limited. During the feature selection process, the subset of features that are most relevant and least redundant is chosen. In recent years, it has become clear that, in addition to relevance and redundancy, features' complementarity must be considered. Informally, if the features are weak predictors of the target variable separately and strong predictors when combined, then they are complementary. It is demonstrated in this paper that the synergistic effect of complementary features mutually amplifying each other in the construction of two-tier decision trees can be interfered with by another feature, resulting in a decrease in performance. It is demonstrated using cross-validation on both synthetic and real datasets, regression and classification, that removing or eliminating the interfering feature can improve performance by up to 24 times. It has also been discovered that the lesser the domain is learned, the greater the increase in performance. More formally, it is demonstrated that there is a statistically significant negative rank correlation between performance on the dataset prior to the elimination of the interfering feature and performance growth after the elimination of the interfering feature. It is concluded that this broadens the scope of feature selection methods for cases where data and computational resources are sufficient.
\end{abstract}

\section{Introduction}

Dimensionality reduction is now sometimes recommended before training a model in a machine learning pipeline. As will be demonstrated below, there is a general consensus in the literature that dimensionality reduction should be considered if there are few samples in the dataset or insufficient computational resources. Otherwise, there's no need to waste time on it. This paper is demonstrated that this is not the case. Even if we have a sufficient number of samples in the dataset (i.e. the number of rows is much greater than the number of columns) and sufficient computational resources, dimensionality reduction can still be useful because we can lose a lot of accuracy if we don't. This does not happen very often, but when it does, the explanatory power may be reduced several times without the use of dimensionality reduction. This is due to the phenomenon of interference with complementarity of the features, as demonstrated in this paper. 

The third section explores the well-known phenomenon of feature complementarity and how it is accounted for in various feature selection methods. The fourth section examines the new phenomenon of interfering with features' complementarity and provides illustrative, clear examples of this phenomenon.  The fifth section uses a synthetic dataset to demonstrate this phenomenon.

The sixth section describes an experiment and its results, which demonstrate the loss in explanatory power that can occur as a result of this phenomenon on real datasets. It was also discovered that the effect of loss of explanatory power becomes stronger when the domain is less learned.

\section{When is it recommended to use dimensional reduction?}

Many seminal reviews \cite{2011, 2017}, as well as the textbook \cite{2020textbook}, argue that the main reasons to use feature selection are (I) to reduce computational and storage costs, (II) to create a cleaner, more interpretable, compact model, and (III) to improve performance metrics in general, and learning accuracy in particular. However, how does dimensionality reduction in general, and feature selection in particular, impact accuracy?

The reasons why irrelevant and/or redundant features can cause a decrease in accuracy are almost never given, but the logical chain is sometimes explained: "Also, with a large number of features, learning models tend to overfit, which may cause performance degradation on unseen data." \cite{2017} Overfitting is thus the only reason for possible accuracy degradation with sufficient time and computational resources when using superfluous features. As a result, it appears plausible that features are not superfluous in the absence of overfitting under conditions of sufficient time and computational resources: in an ideal situation, adding features will not reduce accuracy, but it may not lead to improved accuracy.

The latter thesis is almost never explicitly stated, presumably because it is assumed to be self-evident.
Sometimes, however, this idea is expressed \cite{2011} (in a slightly distorted form): "In theory, increasing the size of the feature vector is expected to provide more discriminating power."

As a result, if we have a dataset with a negligible amount of overfitting and enough time and computational resources, we should avoid using dimensionality reduction. For an engineer, this means that investing time in learning and implementing dimensionality reduction methods is not worthwhile in this case. 

However, we will demonstrate that this is incorrect. And that not using feature selection can sometimes result in a significant loss of accuracy, even if we have a dataset with orders of magnitude more samples than features (so overfitting is insignificant) and a sufficient amount of time and computational resources.

\section{Complementarity of the features}

It is quite common in machine learning to discover that two features can predict the value of the target variable very poorly individually, but very well together. This feature property is known as complementarity. For a long time, it has been recognized that features can be complementary to one another, and that complementarity as such is a new systemic quality that cannot be completely reduced to relevance and redundancy \cite{1997, 2003}. 

However, only a few decades later, algorithms for feature selection that take complementarity into account and thus produce better results began to appear \cite{2008}. In recent years, the understanding that feature selection should be based not on the dyad of relevance -- redundancy, as it was before \cite{2004}, but on the triad of relevance -- redundancy -- complementarity has grown and been finally explicated \cite{2016, 2018, 2020}.

Despite this, the consideration of complementarity is still reflected rather incompletely and briefly in the major reviews on feature selection, sometimes even more so in the earlier reviews \cite{2011} than in the later ones \cite{2017}. 

The more commonly used theoretical-informational formalization of the complementarity phenomenon is not the only one that can be used. Formalizations are also achieved through rank correlations \cite{2020mlsd, 2020rsci} and the construction of two-tier decision trees \cite{2021,2021r}.

\section{Interfering with features' complementarity}

Furthermore, we can show that in some cases, formalizing the phenomenon of complementarity through the construction of a two-tier tree is preferable to theoretical-informational formalizations. Particularly when a "greedy" algorithm is used to build a model from selected features.

In this case, a number of new phenomena emerge "at the intersection" of feature complementarity and "greediness" of the model training algorithm, which are absent when constructing a model using a brute-force approach. The ability to interfere with the effect of mutual feature amplification by some third feature, which we will call the interfering one, is one of these new phenomena. 

As a result, removing the interfering feature improves the trained model's accuracy. Adding a feature to the data, on the other hand, can reduce accuracy if the added feature turns out to be a interfering feature. This is somewhat counterintuitive, because it does not happen when full brute-force algorithms are used.

If we add a new feature to the data, for example, the linear regression model without cross-validation can either become more accurate or remain the same accuracy if the added feature turns out to be completely irrelevant or uninformative. However, adding a feature cannot make a linear regression model less accurate.

And when using "greedy" training algorithms, accuracy may suffer, which is why it makes sense to consider the quartet "relevance -- redundancy -- complementarity -- interfering with complementarity" rather than the triad described above. 

\section{Complementary features on synthetic dataset}

\label{tab1}
\begin{table}[ht]
\begin{center}
\caption{Synthetic dataset for testing complementary features}
\begin{tabular}{|c|c|c|c|c|}
\hline
Number of&\multicolumn{3}{|c|}{Features}& Target \\
\cline{2-4} 
Sample & $f_1$& $f_2$& s& Value \\
\hline
0&	0.0&	0.0&	0.0&	100.0 \\
\hline
1&	0.0&	1.0&	0.0&	20.0  \\
\hline
2&	1.0&	0.0&	0.0&	20.0  \\
\hline
3&	1.0&	1.0&	0.0&	100.0 \\
\hline
4&	0.0&	0.0&	7.0&	96.0  \\
\hline
5&	0.0&	1.0&	7.0&	16.0  \\
\hline
6&	1.0&	0.0&	7.0&	16.0  \\
\hline
7&	1.0&	1.0&	7.0&	96.0  \\
\hline
8&	0.0&	0.0&	14.0&	92.0  \\
\hline
9&	0.0&	1.0&	14.0&	12.0  \\
\hline
10&	1.0&	0.0&	14.0&	12.0  \\
\hline
11&	1.0&	1.0&	14.0&	92.0  \\
\hline
12&	0.0&	0.0&	21.0&	88.0  \\
\hline
13&	0.0&	1.0&	21.0&	8.0   \\
\hline
14&	1.0&	0.0&	21.0&	8.0   \\
\hline
15&	1.0&	1.0&	21.0&	88.0  \\
\hline
16&	0.0&	0.0&	28.0&	84.0  \\
\hline
17&	0.0&	1.0&	28.0&	4.0   \\
\hline
18&	1.0&	0.0&	28.0&	4.0   \\
\hline
19&	1.0&	1.0&	28.0&	84.0  \\
\hline
\end{tabular}
\end{center}
\end{table}

Let $\mathscr{F}$ -- finite set of all features, $\mathscr{F} = \{f_1, ..., f_k \}, k\in\mathbb{N}$. Let $F \subset \mathscr{F}$ -- a subset of the set of all features. Let us assume that $t(F)$ -- some performance score for two-tier decision tree built on features $F$ in greedy fashion.

So, let's $f_1, f_2 \in \mathscr{F}$. Then if $t(\{f_1, f_2\}) > \mathrm{max}(t(\{f_1\}), t(\{f_2\}))$, we will say that $f_1, f_2$ — pair of complementary features.

Let's assume that  $f_1, f_2$ — pair of complementary features and $f_1, f_2, s \in \mathscr{F}$. Then if $t(\{f_1, f_2\}) > t(\{f_1, f_2, s\})$, we will say that $f_1, f_2, s$ is triple with interference where $s$ -- interfering feature and $f_1, f_2$ -- pair of features, the complementarity of which has been interfered with. 

Let's assume that $\{f_1, f_2, s\}$ -- triple with interference. Then define the interfering coefficient $S$ as 

\[ 
S = \frac{t(\{f_1, f_2\})}{t(\{f_1, f_2, s\})} 
\]

In Table 1 is presented synthetic dataset for illustrating phenomemon of interfering with features' complementary. This dataset has 19 samples, 3 features and target value. We try to build the two-tier regression decision tree using CART procedure. 

So, the CART procedure implemented in the library scikit-learn (version 1.0.2) applied to this dataset will give the following decision tree (fig. 1). This tree can only explain a small fraction of the variance of the explanatory variable: $t(\{f_1, f_2, s\}) = 0.0186$. 

If we eliminate the $s$ feature from the dataset and apply the same CART procedure, we get a completely different decision tree (fig. 2) that can explain a significantly larger fraction of the variance: $t(\{f_1, f_2\}) = 0.9804$. So we can conclude that for this case interfering coefficient $S = t(\{f_1, f_2\}) / t(\{f_1, f_2, s\}) \approx 0.9804 / 0.0186 \approx 52.63$.

\begin{figure}[htbp]
\centerline{\includegraphics[scale=0.8]{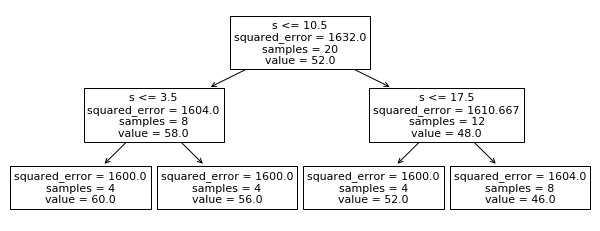}}
\caption{Decision tree building without elimination any features}
\label{fig1}
\end{figure}

\begin{figure}[htbp]
\centerline{\includegraphics[scale=0.8]{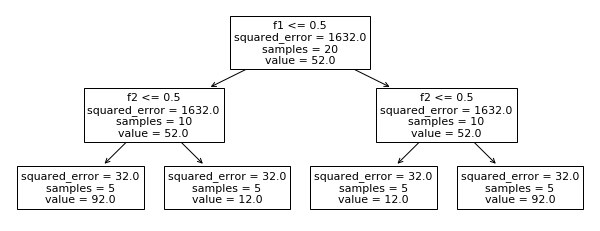}}
\caption{Decision tree building on two complementary features}
\label{fig2}
\end{figure}

So we can see that the $s$ feature is a relatively strong predictor of the target variable, while the $f_1$ and $f_2$ features are weak predictors separately, so if the $s$ feature is not excluded from the dataset, it will "overshadow", interfere with $f_1$ and $f_2$ features and not let any of them appear in the decision tree. On the contrary, if $s$ is excluded from the dataset, it turns out that $f_1$ and $f_2$ together turn out to be able to significantly increase the fraction of the explained variance -- $52.63$ times.

\section{Description of experiments and main results}

We demonstrated the phenomenon of interfering with complementarity on a synthetic dataset without cross-validation in the previous section. The goal of this section is to demonstrate and evaluate this phenomenon on real datasets using cross-validation. To do so, we must first describe the datasets that will be used. The performance metric for classification datasets is then described. The interfering coefficient with cross-validation is then defined.

So, consider the well-known public dataset, the so-called "Boston dataset" (BOS) \cite{BOS}. It contains 506 samples and 14 features, including the target variable, the median price of a house in one of Boston's 506 neighborhoods. Let’s take a subset of this dataset. We leave only the samples corresponding to neighborhoods with houses with a "average" number of rooms in the dataset: that is, when $5 \leq RM < 6.7$. This condition is met by 379 samples.

The BOS dataset is an example of a regression dataset. Let's take a look at two more classification datasets. The first will contain information aimed at predicting the patient's presence or absence of diabetes — DIA \cite{DIA}. The second dataset includes information aimed at predicting whether or not the patient has an arrhythmia — ARR \cite{ARR}.

Let's go over the performance metrics we employ. For regression tasks, we use a fraction of the explained variance as the performance metric. It can take values ranging from 0 to 1. It would seem natural to use accuracy as a metric for classification tasks, but it should be normalized to values ranging from 0 to 1 to be comparable with the results of regression tasks.

Let's take a look at the classification task. If we don't have any features, we can get the best results by always selecting the most frequent class. So, with any informative feature, we can hope for a better results.

Assume we have a binary classification task, $N$ samples in the dataset, and $M$ first class samples, where $1 \leq M < N$. As a result, we have $N - M$ samples of second class. So we can define the minimum performance we expect from any feature.

\[
A^{\mathrm{norm}}_{\mathrm{min}} = \frac{\mathrm{max}(M; N - M)}{N}
\]

Assume that accuracy $A$ is a proportion of correct answers provided by the trained model. Let us now define normalized accuracy $A^{\mathrm{norm}}$ as

\[
A^{\mathrm{norm}} = \frac{A - A^{\mathrm{norm}}_{\mathrm{min}}}{1 - A^{\mathrm{norm}}_{\mathrm{min}}}
\]

Let us now briefly describe the experiment's plan. First, we identify statistically significant triples with interference for each dataset. We discovered 161 such triples for ARR, 11 triples for BOS, and one triplet for DIA. There are 174 triples with interference in total. We know which pair of complementary features is being interfered with which feature for each triple; how accurate the model is on the three features and the two complementary features; and, consequently, what the interfering coefficient is.

Assume that subset of features $\{f_1, f_2, s\}$ is a triple with interference as it was defined in the above section, where $f_1, f_2$ is a pair of complementary features and $s$ -- interfering feature. Let's divide the dataset $D$ into training and test samples with random seed $r_j$, where $j = 1,.., R^D$. Number of such different division $R^D$ is vary from one dataset to another, for example, we use $R^{\mathrm{BOS}} = R^{\mathrm{ARR}} = 1000$ and $R^{\mathrm{DIA}} = 10000$. 

Let's denote $t^j(F)$ -- performance score by two-tier decision tree built on features' subset $F$ on test sample of division dataset $D$ with random seed $r_j$. Then we can get two sets of performance scores -- $\{t^j(\{f_1, f_2, s\})\}$ for the trees with interfering effect and $\{t^j(\{f_1, f_2\})\}$ for the tree without interfering effect after eliminating interfering feature.

So first set of performance scores $\{t^j(\{f_1, f_2, s\})\}$ allows us to evalute confident interval $t^{\mathrm{inter}}_{\mathrm{ci}}$ to the mean value of performance score for building tree with interfering effect $t^{\mathrm{inter}}_{\mathrm{mean}}$. We will use $p$-value$ = 0.05$ in concrete calculations.

\[
t_{\mathrm{ci}}^{\mathrm{inter}} = \Big( t_{\mathrm{min}}(\{f_1, f_2, s\}); t_{\mathrm{max}}(\{f_1, f_2, s\}) \Big)
\]

\[
t_{\mathrm{mean}}^{\mathrm{inter}} = \frac{t_{\mathrm{min}}(\{f_1, f_2, s\}) + t_{\mathrm{max}}(\{f_1, f_2, s\})}{2} 
\]

Using set of performance score $\{t^j(\{f_1, f_2\})\}$ we can evaluate a confident interval $t_{\mathrm{ci}}^{\mathrm{elim}}$ and the mean value of performance score $t_{\mathrm{mean}}^{\mathrm{elim}}$ for the complementary pair of features which will be after eliminating interfering feature from the triple with interference.

\[
t_{\mathrm{ci}}^{\mathrm{elim}} = \Big( t_{\mathrm{min}}(\{f_1, f_2\}); t_{\mathrm{max}}(\{f_1, f_2\}) \Big)
\]

\[
t_{\mathrm{mean}}^{\mathrm{elim}} = \frac{t_{\mathrm{min}}(\{f_1, f_2\}) + t_{\mathrm{max}}(\{f_1, f_2\})}{2} 
\]

Let's determine the minimum value for the interfering coefficient using cross-validation:

\[
S_{\mathrm{min}}^{\mathrm{cv}} = \frac{t_{\mathrm{min}}(\{f_1, f_2\})}{t_{\mathrm{max}}(\{f_1, f_2, s\})} 
\]

Let's determine the maximum value for the interfering coefficient using cross-validation:

\[
S_{\mathrm{max}}^{\mathrm{cv}} = \frac{t_{\mathrm{max}}(\{f_1, f_2\})}{t_{\mathrm{min}}(\{f_1, f_2, s\})} 
\]

And then let's define the interfering coefficient with cross-validation:

\[
S^{\mathrm{cv}} = \frac{S_{\mathrm{min}}^{\mathrm{cv}} + S_{\mathrm{max}}^{\mathrm{cv}}}{2}
\]

Let's note that following this definition

\[
S^{\mathrm{cv}} \neq \frac{t_{\mathrm{mean}}^{\mathrm{elim}}}{t_{\mathrm{mean}}^{\mathrm{inter}}}
\]

For example, on DIA dataset with $R^{\mathrm{DIA}} = 10000$ we can get that $t_{\mathrm{ci}}^{\mathrm{inter}} = (1.714 \cdot 10^{-3}; 5.104 \cdot 10^{-3})$. And $t_{\mathrm{ci}}^{\mathrm{elim}} = (61.018 \cdot 10^{-3}; 64.415 \cdot 10^{-3})$. Therefore, $t_{\mathrm{mean}}^{\mathrm{inter}} = 3.409 \cdot 10^{-3}$, $t_{\mathrm{mean}}^{\mathrm{elim}} = 62.716	 \cdot 10^{-3}$. So, we can say that $t^{\mathrm{inter}} = (3.409 \pm 1.695) \cdot 10^{-3}$ and $t^{\mathrm{elim}} = (62.716 \pm 1.699) \cdot 10^{-3}$.

That means $S^{\mathrm{cv}}_{\mathrm{min}} = \frac{61.018 \cdot 10^{-3}}{5.104 \cdot 10^{-3}} \approx 11.955$. And $S^{\mathrm{cv}}_{\mathrm{max}} = \frac{64.415 \cdot 10^{-3}}{1.714 \cdot 10^{-3}} \approx 37.582$. So $S^{\mathrm{cv}} = \frac{11.955 + 37.582}{2} \approx 24.77$. 

Note, that $t_{\mathrm{mean}}^{\mathrm{elim}}/t_{\mathrm{mean}}^{\mathrm{inter}} = 62.716 \cdot 10^{-3} / 3.409 \cdot 10^{-3} \approx 18.40 \neq 24.77$. But on other datasets, for example, on BOS we can see that $S^{\mathrm{cv}} \approx t_{\mathrm{mean}}^{\mathrm{elim}}/t_{\mathrm{mean}}^{\mathrm{inter}}$.

Now we can gathered together all the information about triples with interference on BOS dataset in Table 2.

\begin{table}[ht]
\label{tab2}
\caption{Triples with interference on BOS dataset}
\begin{center}
\begin{tabular}{ |c||c|c|c|c|c|c| } 
  \hline
  № & $f_1$ & $f_2$ & $s$ & $S^{\mathrm{cv}}$ & $t^{\mathrm{inter}}_{\mathrm{mean}}$ & $t^{\mathrm{elim}}_{\mathrm{mean}}$ \\
 \hline
 \hline
 1 & INDUS & RM & CRIM & 1.24 & 0.214 & 0.264 \\
 2 & INDUS & DIS & CRIM & 1.25 & 0.213 & 0.265 \\
 3 & CHAS & TAX & CRIM & 2.81 & 0.095 & 0.257 \\
 4 & RM & TAX & CRIM & 1.32 & 0.212 & 0.281 \\
 5 & RM & PTRATIO & CRIM & 1.10 & 0.213 & 0.234 \\
 6 & DIS & TAX & CRIM & 1.24 & 0.212 & 0.261 \\
 7 & DIS & PTRATIO & CRIM & 1.31 & 0.215 & 0.280 \\
 8 & DIS & B & CRIM & 1.24 & 0.210 & 0.258 \\
 9 & INDUS & RM & AGE & 1.65 & 0.161 & 0.264 \\
10 & CHAS & TAX & AGE & 1.24 & 0.209 & 0.257 \\
11 & RM & RAD & AGE & 1.43 & 0.154 & 0.220 \\
12 & RM & TAX & AGE & 1.57 & 0.180 & 0.281 \\
 \hline
\end{tabular}
\end{center}
\end{table}

And after that we can gathered together all the information about triples with interference on DIA dataset in Table 3.

\begin{table}[htbp]
\label{tab3}
\caption{Triples with interference on DIA dataset}
\begin{center}
\begin{tabular}{ |c||c|c|c|c|c|c| } 
  \hline
  № & $f_1$ & $f_2$ & $s$ & $S^{\mathrm{cv}}$ & $t^{\mathrm{inter}}_{\mathrm{mean}}$ & $t^{\mathrm{elim}}_{\mathrm{mean}}$ \\
 \hline
 \hline
 1 & Preg-s & DiabPedigree & Age & 24.77 & 0.00341 & 0.06271 \\
 \hline
\end{tabular}
\end{center}
\end{table}

This table with BOS results is a good example to explain how we evalute connection between $t^{\mathrm{inter}}_{\mathrm{mean}}$ and $S^{\mathrm{cv}}$. We calculate Spearman's rank correlation between $t^{\mathrm{inter}}_{\mathrm{mean}}$ and $S^{\mathrm{cv}}$ for all triples with interference and put in Table 4. We see that there is a statistically significant negative rank correlation between these two variables.

Following that, we can compile all of the information about all of the analyzed datasets in Table 4. We will not publish a special table on the ARR dataset because there are too many triples with interference there, but the fundamental principles remain the same.

Let's describe  information in Table 4 more in detail. Assume that the given dataset $D$ has $Q$ triples with interference, where $Q \in \mathbb{N}$. We can conclude that performance score with cross-validation for $i$'s triples with interference will be $t^{\mathrm{inter}}_{\mathrm{mean}, i}$ before elimination the interfering feature. And confident interval for the interfering coefficient for $i$'s triples with interference will be $(S^{\mathrm{cv}}_{\mathrm{min}, i}; S^{\mathrm{cv}}_{\mathrm{max}, i})$. And the mean of interfering coefficient for $i$'s triples with interference will be $S^{\mathrm{cv}}_i$. Let's denote that minimal value of the $S^{\mathrm{cv}}_i$ across all the triples with interference $i = 1,..,Q$ in $D$ as $S^D_{\mathrm{min}}$:

\[
S^D_{\mathrm{min}} = \min_{i} \{S^{\mathrm{cv}}_i\}
\]

Similarly let's denote

\[
S^D_{\mathrm{max}} = \max_{i} \{S^{\mathrm{cv}}_i\} 
\]
\[
S^D_{\mathrm{median}} = \underset{i}{\mathrm{median}} \{S^{\mathrm{cv}}_i\}
\]

And also we can define median value across all the triples with interference of the average performance score before elimination interfering feature $t^D_{\mathrm{median}}$.

\[
t^D_{\mathrm{median}} = \underset{i}{\mathrm{median}} \{t^{\mathrm{inter}}_{\mathrm{mean}, i}\}
\]

And also assume that

\[
\rho^D, p\mathrm{-value} = \rho ( \langle t^{\mathrm{inter}}_{\mathrm{mean}, i} \rangle; \langle S^{\mathrm{cv}}_i \rangle )
\]

\begin{table}
\begin{center}
\label{tab4}
\caption{Triples with interference on all analyzed datasets}
\begin{tabular}{ |c||c|c|c|c|c|c|c|c|c| } 
  \hline
  № & $D$ & $\#$ of t-s & $D$ type & $S^D_{\mathrm{min}}$ & $S^D_{\mathrm{median}}$ & $S^D_{\mathrm{max}}$ & $t^D_{\mathrm{median}}$ & $\rho^D$ & $p\mathrm{-value}$ \\
 \hline
 \hline
 1 & ARR & 161 & C & 1.09 & 1.23 & 2.52 & 0.2117 & -0.76 & $9.8 \cdot 10^{-32}$ \\ 
 2 & BOS & 12 & R & 1.10 & 1.28 & 2.81 & 0.2108 & -0.60 & $3.9 \cdot 10^{-2}$ \\ 
 3 & DIA & 1 & C& 24.77 & 24.77 & 24.77 & 0.0034 & -- & --\\ 
 4 & ALL & 174 & C/R & 1.09 & 1.24 & 24.77 & 0.2117 & -0.76 & $2.4 \cdot 10^{-34}$ \\ 
 \hline
\end{tabular}
\end{center}
\end{table}

So we can conclude from table 4 that (I) negative rank correlation between $t^{\mathrm{inter}}_{\mathrm{mean}, i}$ and $S^{\mathrm{cv}}_i$ is true for different types of datasets — regression and classification; of different size, and so on. Aside from that, we can see that (II) the lower the initial performance score $t^D_{\mathrm{median}}$, the higher the interfering coefficient $S^D_{\mathrm{median}}$.

\section{Conclusion}
The problem of interfering with complementarity described above changes the whole process of building a two-level decision tree. It was previously believed that dimensionality reduction methods should not be considered when overfitting is not expected and sufficient computational resources are available. Overfitting is not expected if the number of rows in a dataset exceeds the number of columns by one or two orders of magnitude (or more). In other words, if you have enough data and computational resources, spending time and effort on dimensionality reduction methods (feature selection and extraction) is considered overkill.  

But, as it turns out, this is not the case. Even with sufficient data and computational resources, failing to use dimensionality reduction methods when building a two-tier decision tree can result in a significant loss of accuracy when compared to using dimensionality reduction methods. Furthermore, the worse the available features predict the target variable, the greater the loss. In other words, the less 'learned' the subject area, the more important it is to focus on dimensionality reduction methods, even if sufficient data and computational resources are available.

\end{document}